# Multi-Objective Dynamic Programming with Limited Precision[1]

L. Mandow , J.L. Pérez de la Cruz , and N. Pozas

Universidad de Málaga, Andalucía Tech, Departamento de Lenguajes y Ciencias de la Computación, Málaga, España.
*lawrence@lcc.uma.es, perez@lcc.uma.es*

September 15, 2020

**Abstract**

This paper addresses the problem of approximating the set of all solutions for Multi-objective Markov Decision Processes. We show that in the vast majority of interesting cases, the number of solutions is exponential or even infinite. In order to overcome this difficulty we propose to approximate the set of all solutions by means of a limited precision approach based on White's multi-objective value-iteration dynamic programming algorithm. We prove that the number of calculated solutions is tractable and show experimentally that the solutions obtained are a good approximation of the true Pareto front.

*Keywords* — Reinforcement learning, Multi-objective Markov decision processes, Multi-objective dynamic programming

## 1   Introduction

Markov decision processes (MDP) are a well-known conceptual tool useful for modelling sequential decision processes and have been widely used in real-world applications such as adaptive production control (see e. g. Kuhnle et al. (2020)), equipment maintenance (see Barde et al. (2019), Liu et al. (2019)) or robot planning (see Veeramani et al. (2020)), to name a few.

Usual optimization procedures take into account just a scalar value to be maximized. However, in many cases the objective function is more accurately described by a vector (see e. g. Gen and Lin (2014), Zhang and Xu (2017)) and multi-objective optimization must be applied.

By merging both concepts (MDP and multi-objective optimization) we are led to consider Multi-objective Markov decision processes (MOMDPs), that are Markov decision processes in which rewards given to the agent consist of two or more independent values, i. e., rewards are numerical vectors. In recent years there has been a growing interest in considering theory and applications of multi-objective Markov decision processes and multi-objective Reinforcement Learning, e.g. see Roijers and Whiteson (2017), and Drugan et al. (2017).

[1] Funded by the Spanish Government, Agencia Estatal de Investigación (AEI), and European Union, Fondo Europeo de Desarrollo Regional (FEDER), grant TIN2016-80774-R (AEI/FEDER, UE).

As in any multi-objective problem, the solution to a MOMDP is not a singleton but in general a set with many *nondominated* values that is called the *Pareto front*. Two approaches can be followed in order to find the solutions of a MOMDP (Roijers et al. (2013)). The *single-policy* approach (Perny and Weng (2010), Wray et al. (2015)) computes a single optimal policy according to some scalarization of user preferences; of course, the process can be repeated as many times as desired in order to find different solutions. On the other hand, the *multi-policy* approach tries to compute simultaneously all the values in the Pareto front (Van Moffaert and Nowé (2014), Ruiz-Montiel et al. (2017)).

In any case, the computation of the Pareto front must face a difficulty: since the number of values in the front is usually huge (in fact, it can be infinite), the computation can be infeasible. Therefore, recent multi-objective reinforcement learning techniques avoid approximating the full Pareto front, or are tested on limited problem instances, like deterministic domains (see Drugan et al. (2017)). So it is desirable to have a procedure that, while being feasible, provides a "good" approximation of the front. That is the goal of the proposal presented here.

The work of Drugan (2019) acknowledges that multi-objective reinforcement learning methods have had "slow development due to severe computational problems". Incorporating elements already widely used in multi-objective evolutionary computation into reinforcement learning methods, like scalarization, and dimensionality reduction through principal component analysis, is a promising avenue of research to overcome these difficulties (see Giuliani et al. (2014)).

On the other hand, the application of multi-objective evolutionary methods in a multi-policy setting (i.e. to directly approximate the Pareto front of MOMDP solutions) seems to remain largely unexplored. Although computationally efficient in many domains, multi-objective evolutionary techniques generally lack guarantees on the optimality or precision of the solutions found. Zitzler et al. (2003) analyzed different metrics that can be used to assess the performance of multi-objective algorithms against a reference Pareto set. Regrettably, these benchmark references are generally not available even for simple stochastic MOMDPs due to their inherent complexity.

A different way to deal with the exponential nature of MOMDPs, is to use *mixture policies* as proposed by Vamplew et al. (2009): given a set of deterministic policies, one of them is chosen probabilistically at the start of the process, and followed onwards. The initial set of policies could be the set of *supported* solutions, calculated by linear scalarized versions of reinforcement learning algorithms. The mixture policies form their convex hull, and would be therefore nondominated. However, there may be situations where mixture policies are not acceptable, e.g. for ethical reasons (Lizotte et al. (2010)). These considerations lead us to propose in this paper a new approach that approximates deterministic policies without relying on mixtures.

The structure of this paper is as follows: first we define formally the concepts related to MOMDPs and prove that under a very restrictive set of assumtions the number of solutions of a MOMDP is tractable (Proposition 1); but we show that if any of these assumptions does not hold, then that number becomes intractable or even infinite. In the following section we present the basic valueiteration algorithm for solving a MOMDP (Algorithm 1) and the modification that we propose (Algorithm $W_{LP}$, given by equation 1), and prove that algorithm $W_{LP}$ computes a tractable number of approximate solutions. Algorithm $W_{LP}$ is then applied to benchmark problems to check if it can provide a "good" approximation of the true Pareto front. Finally some conclusions are drawn.

## 2  Multi-objective Markov Decision Processes

This section defines multi-objective Markov decision processes (MOMDPs) and their solutions. When possible, notation is consistent with that of Sutton and Barto (2018).

A MOMDP is defined by at least the elements in the tuple $(S,A,p,\vec{r},\gamma)$, where: S is a finite set of states; A($s$) is the finite set of actions available at state $s \in$ S; $p$ is a probability distribution such that $p(s,a,s')$ is the probability of reaching state $s'$ immediately after taking action $a$ in state $s$; $\vec{r}$ is a function such that $\vec{r}(s,a,s') \in \mathrm{IR}^q$ is the reward obtained after taking action $a$ in state $s$ and immediately transitioning to state $s'$; and $\gamma \in (0,1]$ denotes the current value of future rewards. The only apparent difference with scalar finite Markov decision processes (MDPs) (e.g., see Sutton and Barto (2018)) is the use of a vector reward function.

A MOMDP can be *episodic*, if the process always starts at a start state $s_o \in$ S, terminates when reaching any of a set of terminal states $\Gamma \subseteq$ S, and before termination there is always a non-zero probability of eventually reaching some terminal state. Otherwise, the process is *continuing*. A MOMDP can be of *finite-horizon*, if the process terminates after at most a given finite number of actions $n$. We say that such process is $n$-step bounded. Otherwise, it is of *infinite-horizon*.

MDPs are frequently used to model the interaction of an agent with an environment at discrete time steps. We define the goal of the agent as the maximization of the expected accumulated (additive) discounted reward over time.

Let us now define the concepts related to solving a MOMDP. A decision rule $\delta$ is a function that associates each state to an available action. A policy $\pi = (\delta_1,\delta_2,...\delta_i,...)$ is a sequence of decision rules, such that $\delta_i(s)$ determines the action to take if the process is in state $s$ at time step $i$. If there is some decision rule $\delta$ such that for all $i$, $\delta_i = \delta$, then the policy is *stationary*. Otherwise, it is *non-stationary*. We denote by $\pi^n = (\delta_1,...,\delta_n)$, $n \geq 1$ the finite $n$-step subsequence of policy $\pi$.

Let $S_t$ and $R_t$ be two random variables denoting the state, and the vector reward received at time step $t$ respectively. The value $\vec{v}_\pi$ is defined as the expected accumulated discounted reward obtained starting at state $s$ and applying policy $\pi$ (analogously to the scalar case, see Sutton and Barto (2018)),

$$\vec{v}_\pi(s) = \mathbb{E}\Big[\sum_{k=0}^{\infty} \gamma^k R_{t+k+1} \mid S_t = s\Big]$$

For any given policy $\pi$, we denote the values of its $n$-step subpolicies $\pi^n$ as $\vec{v}_\pi^n$.

Vector values define only a partial order *dominance* relation. Given two vectors $\vec{u} = (u_1,...u_q)$, $\vec{v} = (v_1,...,v_q)$, we define the following relations: (a) Dominates or equals, $\vec{u} \succeq \vec{v}$ iff for all $i$, $u_i \geq v_i$; (b) Dominates, $\vec{u} \succ \vec{v}$ iff $\vec{u} \succeq \vec{v}$ and $\vec{u} \neq \vec{v}$; (c) Indifference, $\vec{u} \sim \vec{v}$ iff $\vec{u} \not\succeq \vec{v}$ and $\vec{v} \not\succeq \vec{u}$.

Given a set of vectors $X \subset \mathbb{R}^q$, we define the subset of nondominated, or Pareto-optimal, vectors as $ND(X) = \{\vec{u} \in X \mid \nexists \vec{v} \in X, \vec{v} \succ \vec{u}\}$.

We denote by $V^n(s)$ and $V(s)$ the set of nondominated values of all possible $n$-step policies, and of all possible policies at state $s$ respectively. For an $n$-step bounded process, $V^n(s) = V(s)$ for all $s \in S$. The solution to a MOMDP is given by the $V(s)$ sets of all states.

## 2.1 Combinatorial explosion

In this section we analyze some computational difficulties related to solving MOMDPs.

**Proposition 1** *Let us consider an episodic q-objective MDP with initial state $s_0$ satisfying the following assumptions:*

1. The length of every episode is at most d.

2. Immediate rewards $\vec{r}(s,a,s')$ are integer in every component and every component is bounded by $r_{min}$, $r_{max}$.

3. The MPD is deterministic.

4. Discount rate is $\gamma = 1$.

*Then $|V(s_0)| \leq (R \times d + 1)^{q-1}$, where $R = r_{max} - r_{min}$.*

Proof: given assumptions 2, 3 and 4 the value of a policy is a vector with integer components. Given assumptions 1 and 2 these components lie in the interval [$d \times r_{min}$, $d \times r_{max}$]. The interval can contain at most $R \times d + 1$ integers. So there can be at most $(R \times d+1)^q$ different vectors for the values; however, not all of them can be nondominated. Consider the $q-1$ first components $1,...,q-1$ of the vector. There are at most $(R \times d+1)^{q-1}$ different possibilities for them. For each one, just one vector is nondominated: the one having the greatest value for the $q$-th component. Hence there are at most $(R \times d+1)^{q-1}$ nondominated policy values, q.e.d.

Notice that nothing has been assumed about the number of *policies* (except that it is finite, by assumption 1). So in general there are many more policies (an exponential number of them) than values.

Let us consider the graph of figure 1 (adapted from Hansen (1980)) with rewards $\vec{r}(s_i,a_1,s_{i+1}) = (0,1)$ and $\vec{r}(s_i,a_2,s_{i+1}) = (1,0)$. It represents an instance (for $d$ = 3) of a family of deterministic MDPs. Assume $\gamma$ = 1. All episodes are of length $d$ and all the assumptions of Proposition 1 hold. And $V(s_0) = ND(V(s_0)) = \{(d,0),(d-1,1),...,(0,d)\}$ and $|V(s_0)| = d + 1$. Notice that the number of policies is $2^d$.

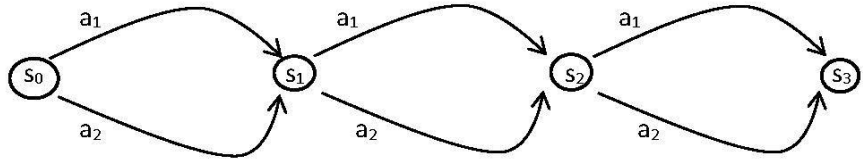

Figure 1: Hansen's graph

We will show now that if *any* of the assumptions is not satisfied, then the number of different nondominated values $|V(s_0)|$ can be also exponential in $d$.

**Nonbounded or noninteger immediate rewards** Let us consider a case where rewards are not bounded and assumption 2 does not hold. Consider for example for all $i$, $\vec{r}(s_i,a_1,s_{i+1}) = (0,2^i)$ and $\vec{r}(s_i,a_2,s_{i+1}) = (2^i,0)$. Then, for depth $d$, for all subset $I \subseteq [1,d]$ and all vector $\vec{v} = (\sum_{i \in I} 2^i, \sum_{i \notin I} 2^i)$ there exists a policy with value $\vec{v}$ for $s_0$, namely the policy that for all state $s_i$ selects $a_1$ if $i \notin I$ and $a_2$ if $i \in I$. Notice that all these values $\vec{v}$ are nondominated. So the number of nondominated values grows exponentially with $d$.

Let us consider now a case with noninteger rewards, for example for all $i$, $\vec{r}(s_i,a_1,s_{i+1}) = (0,1/2^i)$ and $\vec{r}(s_i,a_2,s_{i+1}) = (1/2^i,0)$. Let $b$ be a bit (0 or 1) and $\bar{b}$ its negation (1 or 0). Consider numbers in binary notation. Then, for depth $d$ and for all $\vec{v} = (0.b_1b_2...b_d, 0.\bar{b}_1\bar{b}_2...\bar{b}_d)$ there exists a policy with value $\vec{v}$ for $s_0$, namely the policy that for all state $s_i$ selects $a_1$ if $b_i$ = 0 and $a_2$ if $b_i$ = 1. Notice that all these values $\vec{v}$ are nondominated. So the number of nondominated values grows exponentially with $d$.

**Discount rate** Assume rewards are for all $i$ $\vec{r}(s_i,a_1,s_{i+1}) = (0,1)$ and $\vec{r}(s_i,a_2,s_{i+1}) = (1,0)$, so integer and bounded. But consider a discount rate $\gamma < 1$, for example $\gamma = 1/2$. Then, for depth $d$ and for all $\vec{v} = (0.b_1b_2...b_d, 0.\bar{b}_1\bar{b}_2...\bar{b}_d)$ there exists again a policy (the same defined for noninteger rewards) with value $\vec{v}$ for $s_0$. So the number of nondominated values grows exponentially with $d$.

**Nondeterministic actions** Assume rewards are $\vec{r}(s_i,a_1,s_{i+1}) = (0,1)$ and $\vec{r}(s_i,a_2,s_{i+1}) = (1,0)$, so integer and bounded. Assume also $\gamma = 1$. It is well known that every discounted MDP can be transformed into an equivalent nondiscounted probabilistic MDP (see for instance Puterman (1994), section 5.3). Let us add a final state $g$ to the previous model, and for all $i$ let $p(s_i,a_1,s_{i+1}) = p(s_i,a_2,s_{i+1}) = 1/2$, $p(s_i,a_1,g) = p(s_i,a_2,g) = 1/2$. Then again for depth $d$, for all $\vec{v} = (0.b_1b_2...b_d, 0.\bar{b}_1\bar{b}_2...\bar{b}_d)$ there exists a policy (again the same defined for noninteger rewards) with value $\vec{v}$ for $s_0$, and the number of nondominated values grows exponentially with $d$.

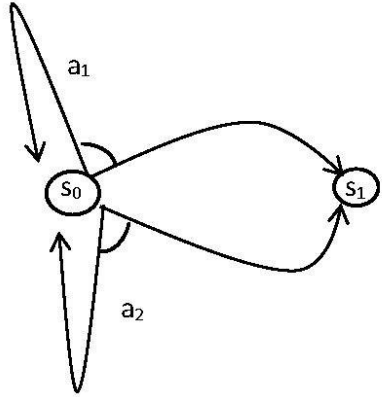

Figure 2: Unbounded episodes

**Cyclical graphs** Let us now withhold assumption 1 (i. e., let us assume that episodes are finite but can be of unbounded length). To avoid the divergence of $\vec{v}$, we must consider that the model is probabilistic, i. e., also abandon assumption 3. Then the situation is even worse: there are MDPs with a finite number of states and an infinite number of nondominated values. This is due to the existence of nonstationary nondominated policies, i. e., policies that take a different action every time a state is reached. For example, let us consider the graph in figure 2. Let $\vec{r}(s_0,a_1,s_0) = (0,1)$, $\vec{r}(s_0,a_2,s_0) = (1,0)$, $\vec{r}(s_0,a_1,s_1) = \vec{r}(s_0,a_2,s_1) = (0,0)$. Let $p(s_0,a_1,s_0) = p(s_0,a_1,s_1) = p(s_0,a_2,s_0) = p(s_0,a_2,s_1) = 1/2$. Then, for all $\vec{v} = (0.b_1b_2...b_i..., 0.\bar{b}_1\bar{b}_2...\bar{b}_i...)$ there exists a nonstationary policy with value $\vec{v}$ for $s_0$. That policy selects for the $i$-th visit to $s_0$ $a_1$ if $b_i = 0$ and $a_2$ if $b_i = 1$. So the number of nondominated values is infinite.

If we consider continuing MDPs -even deterministic- with $\gamma < 1$ the same situation appears: there are nonstationary nondominated policies and an infinite number of values can be obtained. For example, let us consider the graph of figure 3, $\vec{r}(s_0,a_1,s_0) = (0,1)$, $\vec{r}(s_0,a_2,s_0) = (1,0)$, $\gamma = 1/2$. Again for all $\vec{v} = (0.b_1b_2...b_i..., 0.\bar{b}_1\bar{b}_2...\bar{b}_i...)$ the same policy above defined yields the value $\vec{v}$.

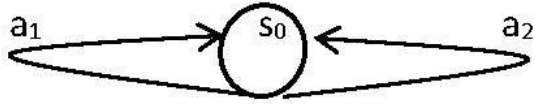

Figure 3: Continuing task

## 3 Algorithms

This section describes two previous multi-objective dynamic programming algorithms, and proposes a tractable modification.

### 3.1 Recursive backwards algorithm

An early extension of dynamic programming to the multi-objective case was described by Daellenbach and De Kluyver (1980). The authors showed that the standard backwards recursive dynamic programming procedure can be applied straightforwardly to problems with directed acyclic networks and additive positive vector costs. The idea easily extends to other decision problems, including stochastic networks. We denote this first basic multi-objective extension of dynamic programming as algorithm *B*.

### 3.2 Vector value iteration algorithm

The work of White (1982) considered the general case, applicable to stochastic cyclic networks, and extended the scalar value-iteration method to multiobjective problems.

Algorithm 1 displays a reformulation of the algorithm described by White. The original formulation considers a single reward associated to a transition $(s,a)$, while we consider that rewards $\vec{r}(s,a,s_i)$ may depend on the reached state $s_i$ as well. Otherwise, both procedures are equivalent.

The procedure is as follows. Initially, the set of nondominated values $V_0(s)$ for each state $s$ is $\{\vec{0}\}$. Line 9 calculates a vector $\vec{s}2 = (s_1,s_2,...s_i,...s_m)$ with all reachable states for the state-action pair $(s,a)$. Let $s2(l) = s_l$ be the *l*-th element in that vector. Then, lines 11 to 13 calculate a list *lv* with all the sets $V_{i-1}(s_l)$ for each state in $\vec{s}2$, i.e. $lv = [V_{i-1}(s_1),...V_{i-1}(s_l),...V_{i-1}(s_m)]$. Function $len(s2)$ returns the length of vector $\vec{s}2$. Next, the cartesian product $P$ of all sets in $lv$ is calculated. Each element in $P$ is of the form $\vec{p} = (\vec{v}_1,... \vec{v}_l,... \vec{v}_m) \mid \forall l\ \vec{v}_l \in V_{i-1}(s_l)$.

---

**Algorithm 1** Algorithm $W$ (adapted from White (1982)).

1: **Input:** size of the state space ($N$); a discount rate ($\gamma$); number of iterations to run ($n$).
2: **Output:** $V_n$, a vector such that $V_n(s) = V^n(s)$
3: $V_0 \leftarrow$ vector(size: $N$, defaultValue: $\{\vec{0}\}$ )
4: **for** $i \in n$ **do**
5:     $V_i \leftarrow$ vector(size: $N$)
6:     **for** $s \in S$ **do**
7:       $T \leftarrow$ vector(size: $|A(s)|$, defaultValue: $\emptyset$)
8:       **for** $a \in A(s)$ **do**
9:         $\vec{s}2 \leftarrow$ reachableStates($s,a$)
10:        $lv \leftarrow$ emptyList()
11:        **for** $j \in$ len($\vec{s}2$) **do**
12:           $lv \leftarrow lv$.append($V_{i-1}(\vec{s}2(j))$)
13:        **end for**
          $P \leftarrow$ cartesianProduct($lv$)
14:        **for** $\vec{p} \in P$ **do**
15:           $T(a) \leftarrow T(a) \cup \left\{\sum_{k \in len(\vec{s}2)}(p(s,a,s2(k)) \times [\vec{r}(s,a,s2(k)) + \gamma\vec{p}(k)]\right\}$
16:        **end for**
17:       **end for**
18:      $V_i(s) \leftarrow$ ND( $\cup_a T(a)$)
19:     **end for**
20:     delete($V_{i-1}$)
21: **end for**
22: return($V_n$)

---

Line 15 calculates a temporal set of updated vector value estimates $T(a)$ for action $a$. Each new estimate is calculated according to the dynamic programming update rule. The calculation takes one vector estimate for each reachable state, and adds the immediate reward and discounted value for each reachable state, weighted by the probability of transition to each such state. All possible estimates are stored in set $T(a)$. Finally, in line 18, the estimates calculated for each action are joined and the non-dominated set is used to update the new value for $V_i(s)$. Maximum memory requirements are determined by the size of $V_i$, $V_{i-1}$ and $T$.

As $n \to \infty$, the values returned by algorithm W converge to V(s) (see White (1982), theorem 1).

Some new difficulties arise in algorithm W, when compared to standard scalar value iteration White (1982):

- After *n* iterations, algorithm W provides $V^n(s)$, i.e. the sets of nondominated values for the set of *n*-step policies (see White (1982), theorem 2). However, for infinite-horizon problems these are only approximations of the V(s) sets. Let us consider two infinite-horizon policies $\pi_1$, and $\pi_2$ with values $\vec{v}_{\pi_1}(s)$, $\vec{v}_{\pi_2}(s)$ respectively at a given state, and all their *n*-step sub-policies $\pi_1^n$ and $\pi_2^n$ with values $\vec{v}_{\pi_1}^n(s)$, $\vec{v}_{\pi_2}^n(s)$. It is possible to have $\vec{v}_{\pi_1}(s) \prec \vec{v}_{\pi_2}(s)$, and at the same time $\vec{v}_{\pi_1}^n(s) \sim \vec{v}_{\pi_2}^n(s)$ for all *n*. In other words, given two infinite-horizon policies that dominate each other, their *n*-step approximations may not dominate each other for any finite value of *n*.

- Nonstationary policies may be nondominated. This is also an important departure from the scalar case, where under reasonable assumptions there is always a stationary optimal policy. White's algorithm converges to the values of nondominated policies, either stationary or nonstationary.

- Finally, if policies with probability mixtures over actions are allowed, these policies may also be nondominated. Therefore, if allowed, these must also be covered in the calculations of the W algorithm.

Despite of its theoretical importance, we are not aware of practical applications of White's algorithm to general MOMDPs. This is not surprising, given the result presented in proposition 1.

## 3.3 Vector value iteration with limited precision

In order to overcome the computational difficulties imposed by proposition 1, we propose a simple modification of algorithm 1. This involves the use of limited precision in the values of vector components calculated by the algorithm. We consider a maximum precision factor. Then, vector components are rounded to the allowable precision in step 18. More precisely, the vectors in the $T(a)$ sets are rounded before being joined. Therefore, all vectors in $V_i$ will be of limited precision.

We denote the *W* algorithm limited to precision ε by $W_{LP}(\varepsilon)$. The new algorithm has a new parameter ε, and line 18 is replaced by,

$$V_i(s) \leftarrow ND(\cup_a \; round(T(a), \epsilon)) \qquad (1)$$

where $round(X, \epsilon)$ returns a set with all vectors in set *X* rounded to ε precision.

**Proposition 2** *Let us consider an episodic q-objective MDP with initial state $s_0$ satisfying the following assumption:*

  1. *Immediate rewards $\vec{r}(s,a,s')$ are bounded by $r_{min}$, $r_{max}$.*

*Then the size of the $V_n(s)$ sets for all states s calculated by $W_{LP}(\varepsilon)$ after n iterations, is bounded by $[(R \times n+1)/\varepsilon]^{q-1}$, where $R = r_{max} - r_{min}$.*

The proof is straightforward from the considerations in the proof of proposition 1, since regardless of the number of possible policies, that is the maximum number of different limited precision vectors that can be possibly generated by the algorithm in *n* iterations.

## 4   Experimental evaluation

This section analyzes the performance of $W_{LP}(\varepsilon)$ on two different MOMDPs: the Stochastic Deep Sea Treasure environment; and the *N*-pyramid environment.

### 4.1   Stochastic Deep Sea Treasure

We first analyze the performance of $W_{LP}(\varepsilon)$ on a modified version of the Deep Sea Treasure (DST) environment from Vamplew et al. (2008). The DST is a deterministic MOMDP defined over a 11 × 10 grid environment, as depicted in figure 4. The true Pareto front for this problem is straightforward. In any case, the environment satisfies all the simplifying assumptions of proposition 1, and therefore can easily be solved with reasonable resources by multi-policy methods.

We introduce the Stochastic DST environment with right-down moves (SDST-RD), as a modified version of the DST task. Our aim is to provide a test environment where the approximations of the Pareto front calculated by multi-policy approaches like $W_{LP}(\varepsilon)$ can be evaluated over a truly stochastic MOMDP. We consider the same grid environment depicted in figure 4. The agent controls a submarine that is placed in the top left cell at the beginning of each episode. The agent can move to adjacent neighboring cells situated down or to its right. Moves outside the grid are not allowed (i.e. the state space has no cycles). When two moves are allowed, the agent actually moves in the specified direction 80% of the time, and in the other allowed direction the remaining 20%. Episodes terminate when the submarine reaches one of the bottom cells. Each bottom cell contains a treasure, whose value is indicated in the cell. The agent has two objectives: maximizing a penalty score (a reward of -1 is received after each move), and maximizing treasure value (the reward value of the treasure is received after reaching a treasure cell).

The state space of the SDST-RD is a directed acyclic graph. Therefore, the exact Pareto front of the start state can be potentially calculated using the recursive backwards algorithm discussed above (we denoted this as algorithm *B*).

This provides a reference for the approximations calculated by $W_{LP}(\varepsilon)$. We are interested in analyzing the performance of the algorithms as a function of problem size. Therefore, we actually consider a set of subproblems obtained form the SDST-RD. The state space of subproblem $i$ comprises all 11 rows, but only the $i$ leftmost columns of the grid shown in figure 4.

The subproblems were solved with algorithm $B$, and with $W_{LP}(\varepsilon)$ with precision $\varepsilon \in \{0.1, 0.05, 0.02, 0.01, 0.001\}$. Smaller values provide higher precision and better approximations. The discount rate was set to $\gamma = 1.0$. With $W_{LP}$, the number of iterations for each subproblem was set to the distance of the most distant treasure.

Figure 5 displays the exact Pareto front for subproblems 1 to 6 as calculated by algorithm B. Larger subproblems exceeded a 96 hour time limit. In general, the Pareto fronts present a non-convex nature, like in the standard deterministic DST. However, the number of nondominated vectors in $V(s_0)$ is much larger (in the standard DST, subproblem $i$ has just $i$ nondominated solution vectors).

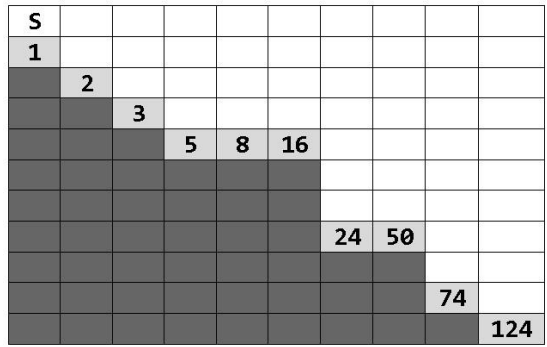

Figure 4: Deep Sea Treasure environment from Vamplew et al. (2008). Start cell is denoted by 'S'.

|        | Algorithm |       |       |       |       |       |
|--------|-----------|-------|-------|-------|-------|-------|
|        | $B$       | $W_{LP}$ | | | | |
| Subpr. |           | 0.001 | 0.01  | 0.02  | 0.05  | 0.1   |
| 1      | 1         | 1     | 1     | 1     | 1     | 1     |
| 2      | 2         | 2     | 2     | 2     | 2     | 2     |
| 3      | 6         | 6     | 6     | 6     | 6     | 5     |
| 4      | 56        | 56    | 45    | 34    | 24    | 15    |
| 5      | 3542      | 1152  | 182   | 107   | 49    | 29    |
| 6      | 34243     | 1923  | 238   | 143   | 58    | 36    |
| 7      | -         | -     | 679   | 344   | 137   | 69    |
| 8      | -         | -     | 602   | 316   | 137   | 72    |
| 9      | -         | -     | -     | 423   | 181   | 94    |
| 10     | -         | -     | -     | 491   | 208   | 108   |

Table 1: Cardinality of $V(s_0)$ for the SDST-RD subproblems.

Only precisions of $\varepsilon \geq 0.02$ were able to solve all 10 subproblems within the time limit. Figure 6 shows the best Pareto frontier approximation obtained for subproblems 7 to 10 by $W_{LP}$ within the time limit.

The cardinality of $V(s_0)$ for each subproblem is shown in table 1. These figures illustrate the combinatorial explosion frequently experienced in stochastic MOMDP even for acyclic state spaces.

Figure 7 shows the maximum number of vectors stored for each subproblem by each algorithm (in logarithmic scale). As expected, the requirements of the exact algorithm grow much faster than any of the approximations. $W_{LP}$ achieves better approximations, and in consequence larger vector sets, with increasing precision (smaller values of $\varepsilon$).

Finally, we briefly examine the quality of the approximations. Figure 8 shows the exact Pareto front and the approximations obtained for subproblem 6. In general, all approximations lie very close, and are difficult to distinguish visually from each other in objective space. Results for other subproblems (omitted for brevity) show a similar pattern. Figure 9 zooms into a particular region of the Pareto front, to display values obtained by different precisions. In general, the approximations are evenly distributed across the true Pareto front, and reasonably close according to the precision values. Table 2 shows the hypervolume of the $V(s_0)$ sets returned by each algorithm for each subproblem. These are calculated relative to the standard reference point (−25,0) used for DST. Differences with the exact Pareto front are almost negligible for $\varepsilon \leq 0.02$.

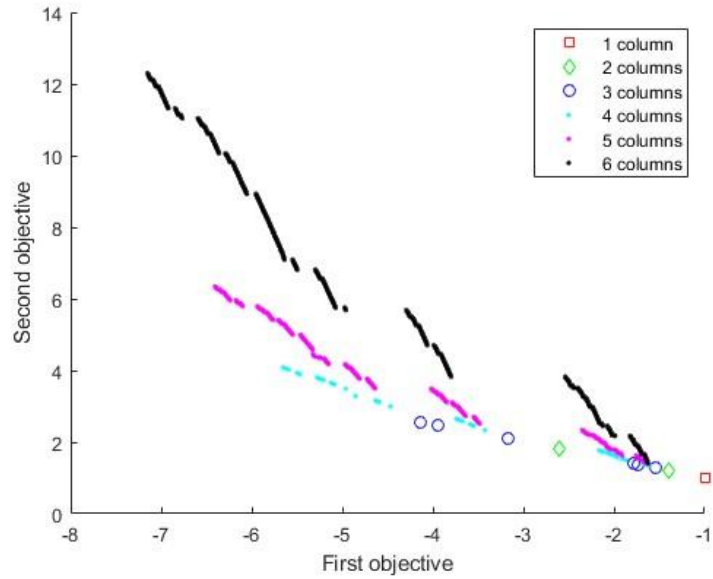

Figure 5: Exact Pareto front for subproblems 1 to 6 of the SDST-RD task calculated with algorithm B.

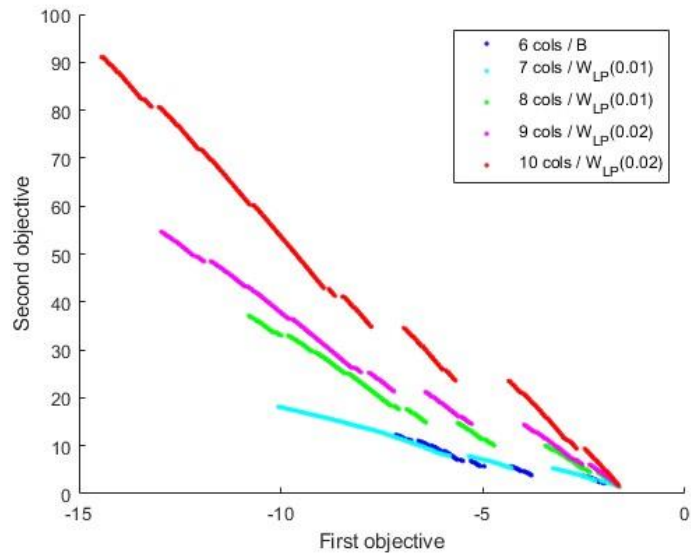

Figure 6: Best approximation of the Pareto frontier obtained for subproblems 7 to 10 of the SDST-RD task with algorithm $W_{LP}$. The exact frontier of subproblem 6 is included as reference.

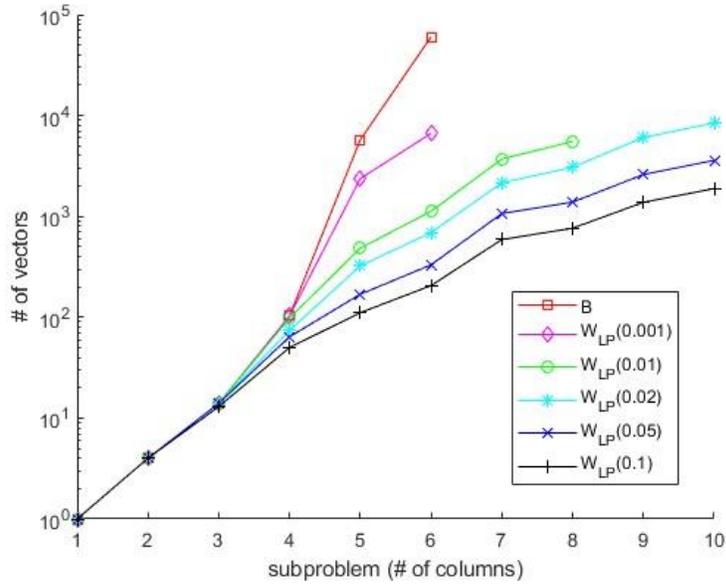

Figure 7: Memory requirements (# of vectors) of each algorithm for the different subproblems of the SDST-RD task (log scale).

## 4.2 *N*-Pyramid

The *N*-pyramid environment presented here is inspired in the pyramid MOMDP described by Van Moffaert and Nowé (2014). More precisely, we define a family of problems of variable size over $N \times N$ grid environments. Each cell in the grid is a different state. Each episode starts always at the start node, situated in the bottom left corner (with coordinates $(x,y) = (1,1)$). The agent can move in any of the four cardinal directions to an adjacent cell, except for moves that would lead the agent outside of the grid, which are not allowed. Transitions are stochastic, with the agent moving in the selected direction with probability 0.95. With probability 0.05 the agent actually moves in a random direction from those available at the state. The episode terminates when the agent reaches one of the cells in the diagonal facing the start state. Figure 10 displays a sample grid environment for $N = 5$. The agent wants to maximize two objectives. For each transition to a non-terminal state, the agent receives a vector reward of $(-1,-1)$. When a transition reaches a terminal cell with coordinates $(x,y)$, the agent receives a vector reward of $(10x, 10y)$. This environment defines a cyclic state space. Therefore, its solution cannot be calculated using the *B* algorithm.

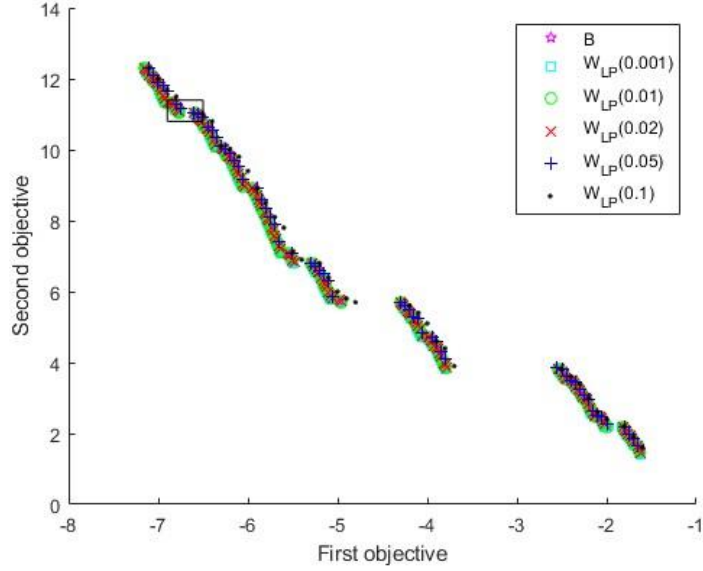

Figure 8: Pareto fronts obtained for SDST-RD subproblem 6. The area in the rectangle is shown magnified in figure 9.

|     | Algorithm | | | | | |
| --- | --- | --- | --- | --- | --- | --- |
|     | $B$ | $W_{LP}$ | | | | |
| Pr. |     | 0.001 | 0.01 | 0.02 | 0.05 | 0.1 |
| 1   | 24.0 | 24.0 | 24.0 | 24.0 | 24.0 | 24.0 |
| 2   | 41.8 | 41.8 | 41.8 | 41.8 | 41.8 | 41.8 |
| 3   | 57.9 | 57.9 | 57.9 | 57.7 | 57.5 | 58.6 |
| 4   | 88.9 | 88.9 | 88.9 | 88.9 | 89.3 | 89.4 |
| 5   | 134.5 | 134.5 | 134.4 | 134.5 | 134.7 | 135.7 |
| 6   | 252.6 | 252.6 | 252.6 | 252.6 | 252.7 | 253.0 |
| 7   | -   | -   | 349.8 | 349.8 | 350.3 | 350.6 |
| 8   | -   | -   | 687.7 | 687.6 | 688.4 | 689.7 |
| 9   | -   | -   | -   | 951.1 | 953.0 | 956.1 |
| 10  | -   | -   | -   | 1513.9 | 1517.9 | 1522.2 |

Table 2: Hypervolumes of $V(s_0)$ for the SDST-RD task.

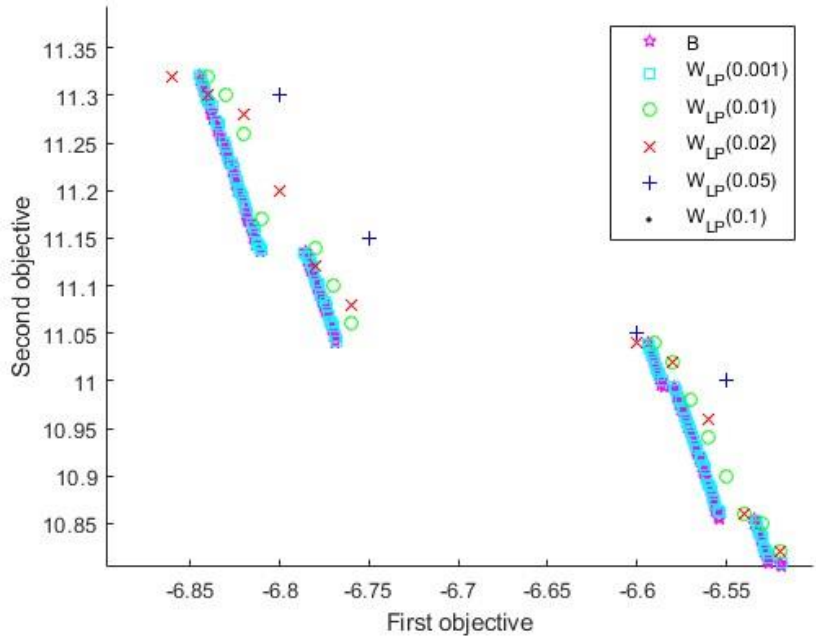

Figure 9: Enlarged portion of objective space from figure 8 (SDST-RD subproblem 6).

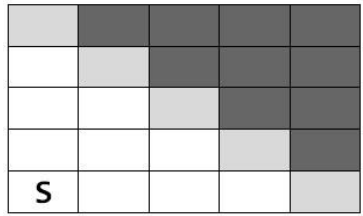

Figure 10: An $N$-pyramid environment for $N$ = 5. The start state is denoted by 'S'. Terminal states are indicated in light grey.

|   | $W_{LP}$ | | | | |
|---|---|---|---|---|---|
| $N$ | 0.005 | 0.01 | 0.05 | 0.1 | 1.0 |
| 2 | 2 | 2 | 2 | 2 | 2 |
| 3 | 137 | 84 | 30 | 20 | 3 |
| 4 | - | - | - | 74 | 8 |
| 5 | - | - | - | - | 19 |

Table 3: Cardinality of $V(s_0)$ for the $N$-pyramid instances.

The N-pyramid environment was solved with $W_{LP}(\varepsilon)$ using $\varepsilon \in \{1.0, 0.1, 0.05, 0.01, 0.005\}$. The discount rate was set to $\gamma = 1.0$, and the number of iterations to $3 \times N$. A twelve hour runtime limit was imposed for all experiments. The largest subproblem solved was $N = 5$, for $\varepsilon = 1.0$.

Figure 11 displays the best frontier approximation for each problem instance, and the corresponding $\varepsilon$ value. Smaller $\varepsilon$ values could not solve the problem within the time limit.

Figure 12 displays the frontiers calculated by the different precision values for the 3-pyramid. This was the largest problem instance solved by all precision values. Figure 13 displays an enlargement of a portion of this frontier. This illustrates the size and shape of the different approximations. As expected, smaller values of $\varepsilon$ produce more densely populated policy values.

Finally, table 3 summarizes the cardinality of the $V(s_0)$ sets calculated for each problem instance by the different precision values, and table 4 the corresponding hypervolumes using the original reference point $(-20,-20)$.

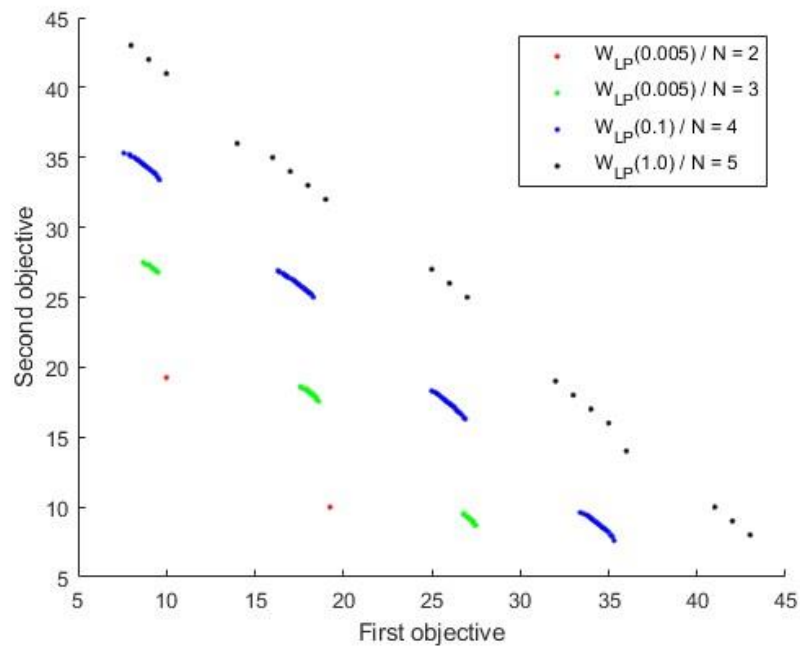

Figure 11: Best approximations obtained for the Pareto fronts for the N-pyramid problem instances.

|   | $W_{LP}$ | | | | |
|---|---------|---------|---------|---------|---------|
| N | 0.005 | 0.01 | 0.05 | 0.1 | 1.0 |
| 2 | 766.25 | 766.25 | 766.25 | 764.00 | 755.00 |
| 3 | 1201.16 | 1200.58 | 1199.40 | 1201.51 | 1180.00 |
| 4 | - | - | - | 1686.61 | 1679.00 |
| 5 | - | - | - | - | 2255.00 |

Table 4: Hypervolumes of $V(s_0)$ for the $N$-pyramid task.

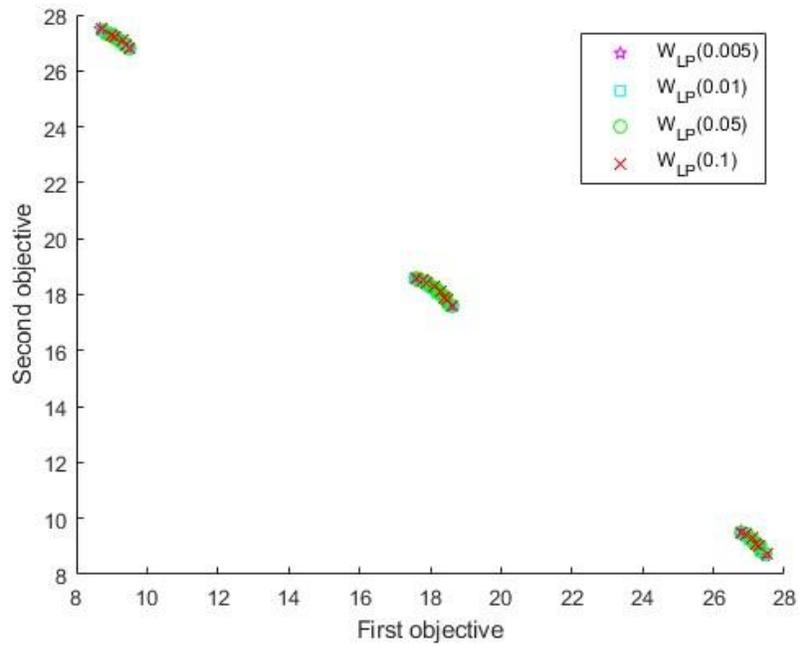

Figure 12: Approximated Pareto fronts for the 3-pyramid instance. An enlarged portion is displayed in figure 13.

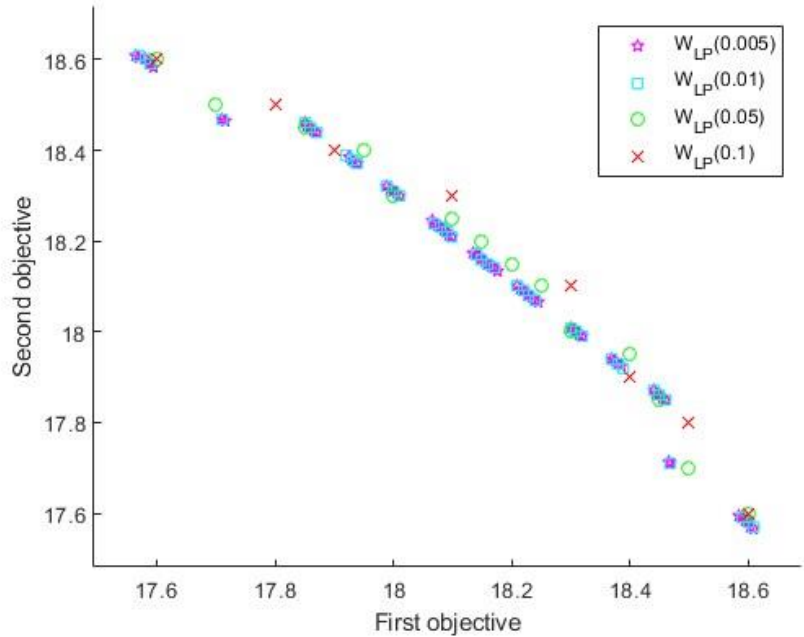

Figure 13: Enlarged portion of objective space from figure 12 (3-pyramid instance).

## 5 Conclusions and future work

This paper analyzes some practical difficulties that arise in the solution of MOMDPs. We show that the number of nondominated policy values is tractable (for a fixed number of objectives $q$) only under a number of limiting assumptions. If any of these assumptions is violated, then the number of nondominated values becomes intractable or even infinite with problem size in the worst case.

Multi-policy methods for MOMDPs try to approximate the set of all nondominated policy values simultaneously. Previous works have addressed mostly deterministic tasks satisfying the limiting assumptions that guarantee tractability.

We show that, if policy values are restricted to vectors with limited precision, the number of such values becomes tractable. This idea is applied to a variant of the algorithm proposed by White (1982). This new variant has been analyzed over a set of stochastic benchmark problems. Results show that good approximations of the Pareto front can be obtained. To our knowledge, the results reported deal with the hardest stochastic problems solved to date by multi-policy multi-objective dynamic programming algorithms.

Future work includes the application of the limited precision idea to more general value iteration algorithms, and the generation of additional benchmark problems for stochastic MOMDPs. This should support the exploration and evaluation of other approximate solution strategies, like multi-policy multiobjective evolutionary techniques.